%% file: main.tex
\documentclass[10pt,twocolumn,letterpaper]{article}

\usepackage{cvpr}              % To produce the CAMERA-READY version
\input{preamble}
\definecolor{cvprblue}{rgb}{0.21,0.49,0.74}

\usepackage{url}            % simple URL typesetting
\usepackage{booktabs}       % professional-quality tables
\usepackage{amsfonts}       % blackboard math symbols
\usepackage{nicefrac}       % compact symbols for 1/2, etc.
\usepackage{microtype}      % microtypography
\usepackage{xcolor}         % colors
\usepackage{utfsym}

\usepackage{xspace}
\usepackage{amssymb}
\usepackage{amsmath}
\usepackage{wrapfig}
\usepackage{algorithm}
\usepackage{algorithmic}

\usepackage{multicol}
\usepackage{multirow}
\usepackage{makecell}
\usepackage{tabularx}
\usepackage{adjustbox}
\usepackage{enumitem}

\usepackage{pifont}
\usepackage{color}
\usepackage{colortbl}

\definecolor{linkcolor}{RGB}{255,0,0}
\definecolor{urlcolor}{RGB}{255,105,180}
\definecolor{citecolor}{RGB}{66,168,235}
\usepackage[pagebackref,breaklinks=true,colorlinks,citecolor=blue,urlcolor=blue,linkcolor=blue,bookmarks=false]{hyperref}
\AtEndPreamble{
    \usepackage[capitalize]{cleveref}
    \crefname{section}{Sec.}{Secs.}
    \Crefname{section}{Section}{Sections}
    \crefname{table}{Tab.}{Tabs.}
    \Crefname{table}{Table}{Tables}
}
\hypersetup{colorlinks=true,linkcolor=linkcolor,urlcolor=urlcolor,citecolor=blue}

\definecolor{lightgray}{rgb}{0.8, 0.8, 0.8}
\definecolor{lgray}{rgb}{0.66, 0.66, 0.66}

\definecolor{lblu_tab}{RGB}{225, 235, 246}
\definecolor{orange_vitad}{RGB}{222, 131, 68}
\definecolor{blue_vitad}{RGB}{106, 153, 208}
\definecolor{trajectory_green}{RGB}{126, 171, 85}
\definecolor{trajectory_yellow}{RGB}{245, 194, 66}
\definecolor{tab_others}{RGB}{235, 235, 235}
\definecolor{tab_ours}{RGB}{225, 235, 246}

\definecolor{whit_tab}{RGB}{255, 255, 255}
\definecolor{gray_tab}{RGB}{246, 246, 246}
\definecolor{oran_tab}{RGB}{252, 242, 237}
\definecolor{blue_tab}{RGB}{227, 240, 251}

\def\onedot{.\xspace}

\def\ie{\textit{i.e}\onedot}

\newcommand{\wyj}[1]{\textcolor{black}{#1}} % new content

\input{symbols}
\def\paperID{5790} % *** Enter the Paper ID here
\def\confName{CVPR}
\def\confYear{2026}

\title{Collaborative Face Experts Fusion in Video Generation: Boosting Identity Consistency Across Large Face Poses}

\author{%
     Yuji Wang \textsuperscript{1}\thanks{Equal contribution.} \thanks{This work was done during internship at Tencent Youtu Lab.},
    Moran Li \textsuperscript{2,*}, 
    Xiaobin Hu\textsuperscript{3,*},
    Ran Yi \textsuperscript{1}, 
    Jiangning Zhang \textsuperscript{2},
    Chengming Xu \textsuperscript{2}, 
    \\
    Weijian Cao \textsuperscript{2},
    Yabiao Wang\textsuperscript{2},
    Chengjie Wang \textsuperscript{2},
    Lizhuang Ma \textsuperscript{1} \thanks{Corresponding author.}
  \\
  \\
\textsuperscript{1}Shanghai Jiao Tong University,
\textsuperscript{2}Tencent Youtu Lab,
\textsuperscript{3}National University of Singapore \\ \\
  {
    \tt 
    \small
    \{moranli,vtzhang,chengmingxu,weijiancao,caseywang,jasoncjwang\}@tencent.com,
  } 
  \\
  {
    \tt 
    \small
     \{yujiwang,ranyi\}@sjtu.edu.cn, xiaobin.hu@tum.de, ma-lz@cs.sjtu.edu.cn
  }
}

\begin{document}
\maketitle
\input{secs/0_abstract}
\input{secs/1_introduction}
\input{secs/2_related_work}
\input{secs/3_method}

\input{secs/4_dataset}
\input{secs/5_experiment}
\input{secs/6_conclusion}

{
    \small
    \bibliographystyle{ieeenat_fullname}
    \bibliography{main}
}

% \input{secs/7_appendix}


\end{document}

%% file: symbols.tex
\newcommand{\fid}{\mathbf{f}_{\text{id}}}
\newcommand{\fsem}{\mathbf{f}_{\text{sem}}}
\newcommand{\fdet}{\mathbf{f}_{\text{det}}}

\newcommand{\Pid}{P_{\text{id}}}
\newcommand{\Psem}{P_{\text{sem}}}
\newcommand{\Pdet}{P_{\text{det}}}

\newcommand{\Eid}{\mathcal{E}_{\text{id}}}
\newcommand{\Esem}{\mathcal{E}_{\text{sem}}}
\newcommand{\Edet}{\mathcal{E}_{\text{det}}}

\newcommand{\eid}{\mathbf{e}_{\text{id}}}
\newcommand{\esem}{\mathbf{e}_{\text{sem}}}
\newcommand{\edet}{\mathbf{e}_{\text{det}}}

\newcommand{\ourdata}{LaFID} 
\def\method{CoFE}

%% file: secs/0_abstract.tex
\begin{abstract}
Current video generation models struggle with identity preservation under large face poses, primarily facing two challenges: the difficulty in exploring an effective mechanism to integrate identity features into DiT architectures, and the lack of targeted coverage of large face poses in existing open-source video datasets. 
To address these, we present two key innovations. 
First, we propose Collaborative Face Experts Fusion (\method), which dynamically fuses complementary signals from three specialized experts within the DiT backbone: an identity expert that captures cross-pose invariant features, a semantic expert that encodes high-level visual context, and a detail expert that preserves pixel-level attributes such as skin texture and color gradients.
Second, we introduce a data curation pipeline comprising three key components: Face Constraints to ensure diverse large-pose coverage, Identity Consistency to maintain stable identity across frames, and Speech Disambiguation to align textual captions with actual speaking behavior. This pipeline yields \ourdata-180K, a large-scale dataset of pose-annotated video clips designed for identity-preserving video generation.
Experimental results on several benchmarks demonstrate that our approach significantly outperforms state-of-the-art methods in face similarity, FID, and CLIP semantic alignment. \href{https://rain152.github.io/CoFE/}{Project page}.
\end{abstract}
% 更完善的版本
% First, we propose a \emph{Collaborative Face Experts Fusion (\method)} framework that dynamically fuses complementary cues from three specialized experts, each designed to capture distinct but mutually reinforcing aspects of facial attributes. The identity expert captures cross-pose identity-sensitive features, the semantic expert extracts high-level visual semantics, and the detail expert preserves pixel-level attributes (\textit{e.g.,} skin texture, color gradients). 
% Furthermore, to mitigate dataset limitations, we design a data processing pipeline with three  components: \emph{Face Constraints} for diverse large-pose coverage and prominent facial regions, \emph{Identity Consistency} to maintain stable identity features across frames, and \emph{Speech Disambiguation} that explicitly annotates speaking status in captions to align text with actual speech behavior and reduce pose-related ambiguity. Leveraging this pipeline, we curate \ourdata-180K~(Large Face Pose for Identity Preservation Dataset), a large-scale collection of video clips with annotated face poses, mainly derived from open-source video datasets. 

%% file: secs/1_introduction.tex
\section{Introduction}
Video generation has emerged as a critical technology for applications ranging from virtual avatars and interactive storytelling to realistic video editing. 
A critical yet unresolved challenge is maintaining the identity consistency of human characters, as \emph{human visual perception is inherently more sensitive to identity inconsistencies in human portraits than in other subjects, particularly when face poses undergo large variations}. 
Such consistency is essential for narrative coherence in storybook generation, visual realism, and reliable character representation in multi-scene animations, as even subtle identity drift can severely undermine user immersion.
\input{figs/teaser}

Recent Diffusion Transformer~(DiT)-based image-to-video models such as CogVideoX~\cite{yang2024cogvideox} and Wan Model~\cite{wan2025wan} achieve content consistency by injecting first-frame VAE features into the DiT backbone. To better preserve character identity, human-centric approaches like ConsisID~\cite{yuan2025consisid} further inject a  frequency-aware facial representation into DiT. However, both paradigms share critical limitations: they rely on a single face feature extracted from the reference frame and inject the same static face feature uniformly across all timesteps and DiT layers, thereby limiting the model’s ability to capture discriminative facial details. This leads to unstable identity preservation under large face poses, often resulting in distorted character appearances. 
Furthermore, existing human-centric datasets~\cite{wang2024humanvid,li2025openhumanvid} are ill-suited for learning pose-robust identity preservation. They often feature faces of inconsistent quality, lack sufficient dynamic large-angle examples, and seldom ensure cross-frame identity consistency or sufficient facial visibility, hindering the learning of pose-invariant identity representations.
% \wyj{Furthermore, existing human-centric datasets~\cite{wang2024humanvid,li2025openhumanvid} are poorly suited for training under large face poses: they often contain faces of inconsistent quality and have a low proportion of dynamic, large-angle examples. They also rarely ensure cross-frame identity consistency or enforce strict facial visibility, making them inadequate for learning pose-robust identity preservation.}

To address these gaps, we propose Collaborative Face Experts Fusion (\method), a hierarchical feature integration framework that overcomes the limitations of static, single-feature identity injection. Since fixed features from the first frame often fail to provide reliable identity cues under large face poses, \method~instead dynamically fuses complementary signals from three specialized experts via a learnable gating mechanism: \textit{i) Identity Expert} extracts pose-robust identity features for cross-view consistency; \textit{ii) Semantic Expert} captures high-level visual concepts (\textit{e.g.,} hairstyle, expression); and \textit{iii) Detail Expert} preserves pixel-level fidelity (\textit{e.g.,}, skin texture, color gradients). By aligning expert contributions with the functional role of each DiT layer, \method~enables robust feature learning under large face poses.

Complementing our architectural innovation, we introduce \ourdata-180K (\textbf{La}rge \textbf{F}ace Pose for \textbf{Id}entity Preservation Dataset), a curated large-scale collection of video clips annotated with 3D face poses. This dataset specifically targets the scarcity of high-quality, identity-consistent videos under large face poses. \ourdata-180K is built through a three-stage pipeline that directly addresses key data limitations: \emph{Face Constraints} ensure clear facial visibility and diverse large-pose coverage; \emph{Identity Consistency} enforces temporal identity coherence via feature clustering; \wyj{and \emph{Speech Disambiguation} aligns video captions with lip movements by verifying speaking status, ensuring accurate annotations under pose-induced ambiguity. We evaluate our method on two distinct benchmarks: the existing VIP-Test~\cite{vip-200k} for general full-body motion, and our newly introduced \ourdata-Bench, which addresses the underexplored setting of large face poses.}

Our contribution can be summarized as follows:

\begin{itemize}
\item We propose Collaborative Face Experts Fusion (\method), a dynamic, multi-source identity injection module that adaptively fuses three specialized experts across DiT layers for robust identity preservation under large face poses.

\item We introduce \ourdata-180K, a dataset of 180K video clips with 3D face pose annotations, curated specifically for large-pose identity preservation via Face Constraints, Identity Consistency, and Speech Disambiguation to ensure diverse large face poses, stable identity across frames, and accurate alignment between lip movements and captions.

\item \wyj{We establish \ourdata-Bench for large face pose evaluation and show that \method~achieves state-of-the-art results on both \ourdata-Bench and VIP-Test, with strong gains under large face poses.}
\end{itemize}

%% file: figs/teaser.tex
% caption表达委婉一点
\begin{figure}[t]
\centering
  \begin{subfigure}[b]{\linewidth}
    \centering
    \includegraphics[width=0.95\linewidth]{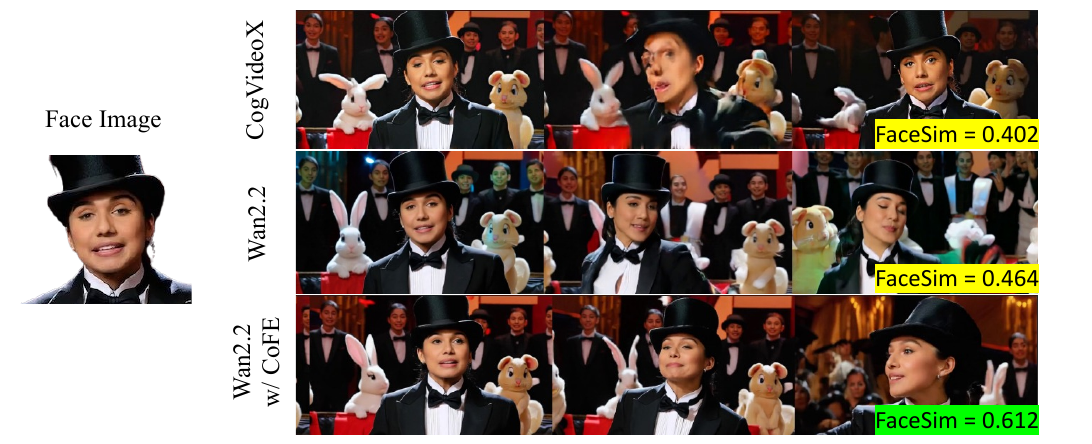}
    \label{subfig:identity_failure}
  \end{subfigure}
  
  \hfill
    \begin{subfigure}[b]{\linewidth}
    \centering
    \includegraphics[width=0.95\linewidth]{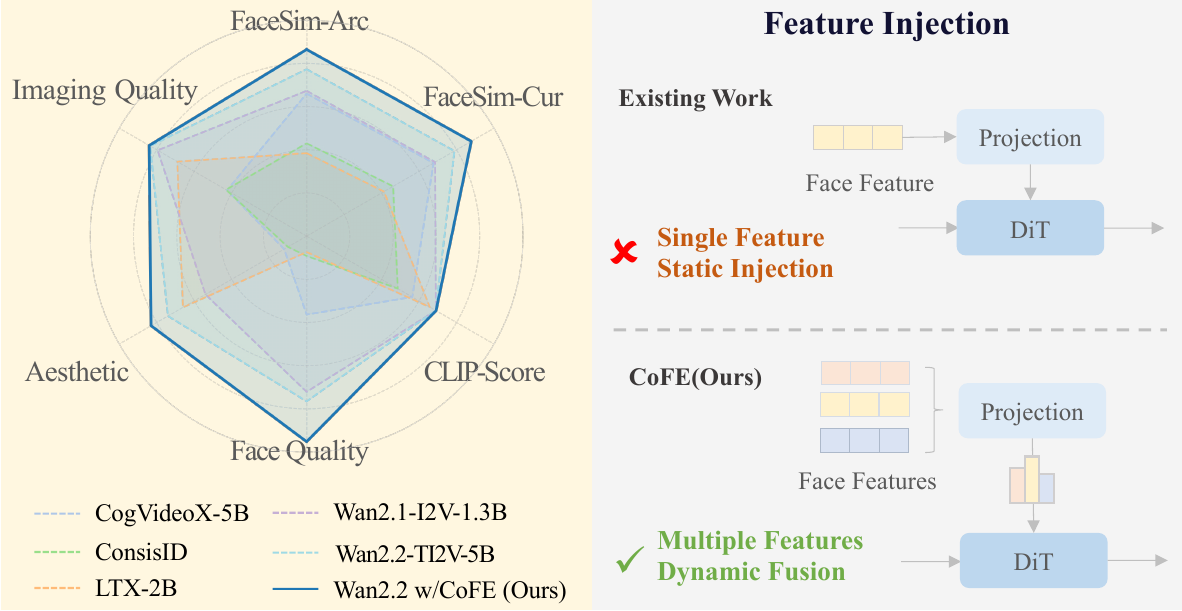}
    \label{subfig:radar}
  \end{subfigure}
  \caption{Top: Identity preservation under large face poses is challenging in current methods, while our approach shows clear gains in face similarity metrics. 
Bottom: With dynamic multi-feature injection, our proposed method achieves the best or competitive performance across all metrics. }
  \vspace{-2mm}
\end{figure}
% The input images are from OpenHumanVid~\cite{li2025openhumanvid}.

%% file: secs/2_related_work.tex
\section{Related Work}
\label{related_work}

\input{figs/pipeline}

\subsection{Identity-Preserving Text-to-Video Generation}
In the identity-preserving text-to-video (IPT2V) task,  early methods used time-consuming fine-tuning for facial identity preservation~\cite{chefer2024still,wu2024motionbooth,wei2024dreamvideo,zhong2025concat,lei2024comprehensive,ma2025controllable}, driving tuning-free approaches to dominance. Pioneering works like ID-Animator~\cite{he2024idanimator} relied on static image-model features, harming dynamic identity consistency.  ConsisID~\cite{yuan2025consisid} injected facial frequency signals but suffered from 2D pose mismatches; FantasyID~\cite{zhang2025fantasyid} added 3D priors but lacked adaptive feature selection. Recent methods like VACE~\cite{jiang2025vace} and SkyReels-A2~\cite{fei2025skyreels} used reference branches but failed to dynamically prioritize cues during large motions. These show I2V needs adaptive fusion of face features aligned with motion, especially large face motions. Thus, we introduce Collaborative Face Experts Fusion (\method), a DiT layer-adaptive gating mechanism that dynamically fuses multiple features to overcome static fusion and enhance large-pose robustness. 

\subsection{Human-Centric Image-to-Video Generation}
Image-to-video (I2V) generation~\cite{guo2024i2v,bharadhwaj2024gen2act,xu2025freevis} struggles to control content reliably with text alone, so recent methods use first-frame conditioning for better controllability. AnimateAnything~\cite{dai2023animateanything} and PixelDance~\cite{zeng2024make} modify U-Nets by concatenating first-frame latent features with input noise to anchor spatial features for temporal consistency. Dynamicrafter~\cite{xing2024dynamicrafter} and Moonshot~\cite{zhang2024moonshot} enhance conditioning with image cross-attention layers, injecting stronger visual cues into diffusion. Recent methods like Wan2.1, Wan2.2~\cite{wan2025wan}, and VACE~\cite{jiang2025vace} support I2V with strong performance but struggle to preserve identity across large face poses. Moreover, IPRO~\cite{shen2025identity} introduces reinforcement learning to enhance training for facial details. 
Generally, identity robustness under large face pose variations remains largely unaddressed in current I2V frameworks.

\subsection{Human-Centric Video Datasets}
Large-scale human-centric video datasets~\cite{sadoughi2015msp,chen2024panda,bain2021frozen} are foundational for human video generation, yet most prioritize body motion or general scenes over facial identity under large poses. For example, TikTok~\cite{jafarian2021learning} focuses on social dance; UCF101~\cite{soomro2012ucf101} and NTU RGB+D~\cite{shahroudy2016ntu} on action recognition; HumanVid~\cite{wang2024humanvid} on synthetic-real blends; \wyj{and OpenHumanVid~\cite{li2025openhumanvid} provides high-resolution videos but lacks large face pose diversity, facial quality filtering, and fine-grained 3D face pose annotations, limiting its suitability for pose-robust identity learning.} To address it, we introduce \ourdata-180K, a dataset constructed by combining OpenHumanVid~\cite{li2025openhumanvid} with carefully selected in-house videos, and curated through a unified pipeline to emphasize diverse large face poses with 3D face pose annotations. 

%% file: figs/pipeline.tex
\begin{figure*}[t]
    \centering
    \includegraphics[width=0.95\linewidth]{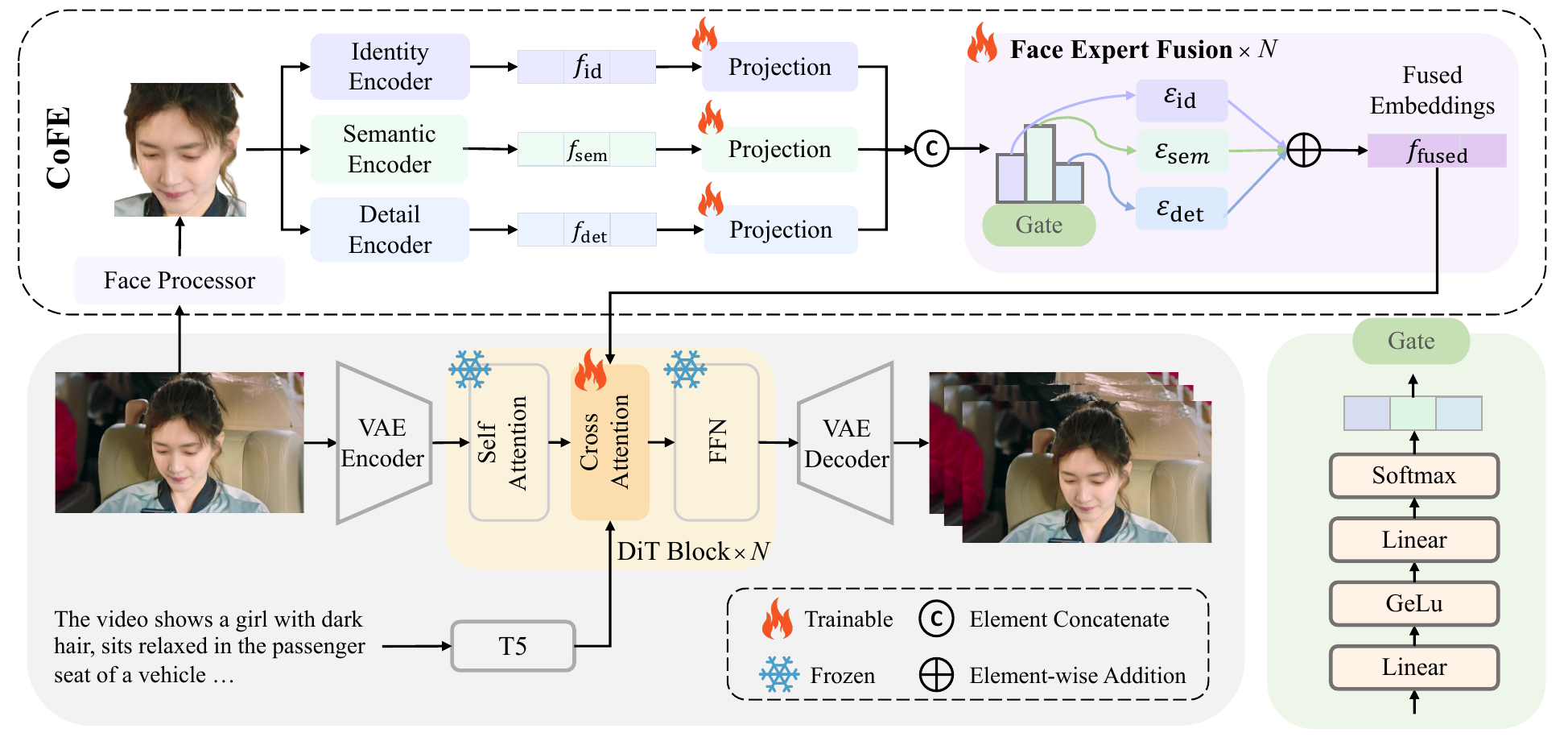}
    \caption{Overview of our pipeline. Our CoFE module comprises globally shared projection layers to align different face features into the feature space of DiT and per-DiT-block expert fusion layers for dynamic feature fusion. Specifically, we employ ArcFace, CLIP, and DINOv2 as our identity, semantic, and detail encoders, respectively. }
    \label{fig:pipeline}
    \vspace{-3mm}
\end{figure*}
% The input images are from OpenHumanVid~\cite{li2025openhumanvid}.

%% file: secs/3_method.tex
\section{Methodology: \method} \label{sec:method}

Current image-to-video models~\cite{yang2024cogvideox,hacohen2024ltx,wan2025wan,jiang2025vace} inject identity using a single-source feature extracted from the first frame, which is then applied statically across all timesteps and network layers, typically via uniform cross-attention. This design suffers from two fundamental limitations. First, the single feature conflates identity, semantics, and details into one representation, lacking specialization. Second, its static injection cannot adapt to pose-induced appearance changes or layer-specific attention demands, leading to severe identity drift under large pose variations. To overcome this, we propose \emph{Collaborative Face Experts Fusion} (\method), a layer-adaptive fusion module that dynamically integrates three specialized facial representations into cross-attention: pose-robust identity features, semantic attributes (\textit{e.g.}, hairstyle, expression), and pixel-level details (\textit{e.g.}, color gradients). An overview of our pipeline is shown in Fig.~\ref{fig:pipeline}.

\input{figs/cross_attn}

\noindent \textbf{Motivation from Cross-Attention Patterns.}
\wyj{To understand how different facial representations influence identity preservation, we analyze cross-attention patterns in the generator. As shown in Figure~\ref{fig:cross_attn}, each representation induces a distinct attention pattern: the Identity Expert (ArcFace~\cite{deng2019arcface}) focuses on pose-invariant landmarks such as the eyes, nose, and mouth to capture structural identity; the Semantic Expert (CLIP~\cite{radford2021learning}) attends to high-level attributes including hairstyle and expression; and the Detail Expert (DINOv2~\cite{oquab2023dinov2}) highlights fine-grained textures and structural edges. These complementary roles indicate that robust identity preservation under large pose variations requires dynamic fusion of all three signals rather than reliance on a single static feature.}

\input{tables/dataset_comp}

\noindent \textbf{Collaborative Face Experts Fusion.} Our goal is to fuse hierarchical face features adaptive to distinct DiT blocks, thereby achieving complementary synergies that enhance identity persistence, semantic coherence, and pixel-level fidelity. To achieve this, we first employ a face preprocessor comprising a face detector~\cite{deng2020retinaface} and a face parsing model~\cite{geitgey2019face} to obtain the cropped and aligned face image from the input image, removing background to focus on the face region. 
These preprocessed faces are then processed by identity, semantic, and detail encoders to generate identity-sensitive features, pixel-level detail features, and high-level semantic features, denoted as $\fid, \fsem, \fdet$, respectively.
% Specifically, we use ArcFace, CLIP, and DINOv2 as the identity, semantic, and detail encoders. 

The proposed \method~module comprises two core components: \textit{i)} Three globally shared projection layers, \ie, $\Pid$, $\Psem$, $\Pdet$, which project $\fid$, $\fsem$, $\fdet$ into a unified feature space compatible with pre-trained DiT blocks, ensuring cross-modality alignment; \wyj{\textit{ii)} Per-DiT-block expert layers, each equipped with three dedicated experts, \ie, $\Eid$, $\Esem$, $\Edet$, and a lightweight gating mechanism that dynamically fuses the projected features. } 
Each expert layer is tailored to the semantic depth of its DiT block, using adaptive gating (via a learnable MLP) to prioritize experts based on the block’s role in denoising processing. This design ensures global feature consistency through shared projection layers and local adaptivity through block-specific expert weighting, enabling end-to-end optimization of cross-dimensional facial representations.
Formally, each expert refines its input feature for the $i$-th DiT block as:
\[
\mathbf{e}^i_\text{attr} = \mathcal{E}^i_\text{attr}(P_\text{attr}(\mathbf{f}_\text{attr})), \text{attr} \in \{\text{id}, \text{sem}, \text{det}\}
\]
Then a learnable gate mechanism computes adaptive weights for these experts based on the concatenated input features $\mathbf{e}^i_{\text{c}} = [\eid^i, \esem^i, \edet^i]$:  
\[
\mathbf{w}^i = \text{Softmax}\left( \mathcal{G}(\mathbf{e}^i_{\text{c}}) \right),
\]
where $\mathcal{G}$ is a linear projection mapping $\mathbf{e}_{\text{c}}$ to weight vector $\mathbf{w}^i = [w^i_{\text{id}}, w^i_{\text{sem}}, w^i_{\text{det}}]$ with elements summing to 1. The final fused face feature, \ie, $\mathbf{f}^i_{\text{fused}}$, can be obtained via:  
\[
\mathbf{f}^i_{\text{fused}} = \mathbf{w}^i \cdot \mathbf{e}^i_c
\]

\noindent\textbf{DiT Feature Injection}
To integrate the fused face features with the DiT architecture, the output of the \method~module, $\mathbf{f}^i_{\text{fused}}$, is injected into the cross-attention layers of the $i$-th DiT block, enabling video latents to learn face-specific features throughout the denoising process. This injection introduces a dedicated facial cross-attention stream that operates in parallel with DiT’s existing context and image cross-attention streams. Each stream contributes key and value vectors: \(\mathbf{k}^i_{\text{ctx}}, \mathbf{v}^i_{\text{ctx}}\) are derived from context, \(\mathbf{k}^i_{\text{img}}, \mathbf{v}^i_{\text{img}}\) are extracted from the first frame to preserve initial visual consistency, and \(\mathbf{k}^i_{\text{fused}}, \mathbf{v}^i_{\text{fused}}\) are projected from \(\mathbf{f}^i_{\text{fused}}\) via learnable matrices to emphasize refined face features. Notably, the gate mechanism in \method~dynamically adjusts the influence of the facial stream, adapting to various demands of different DiT blocks to achieve flexible face feature injection. The aggregated cross-attention output updates video latents as:
\[
\begin{aligned}
\mathbf{o}^i &= \text{attn}^i(\mathbf{q}^i, \mathbf{k}^{i}_{\text{ctx}}, \mathbf{v}^i_{\text{ctx}}) +  \text{attn}^i(\mathbf{q}^i, \mathbf{k}^i_{\text{img}}, \mathbf{v}^i_{\text{img}}) \\
            &\quad + \text{attn}^i(\mathbf{q}^i, \mathbf{k}^i_{\text{fused}}, \mathbf{v}^i_{\text{fused}}),
\end{aligned}
\]
where \(\text{attn}(\cdot)\) denotes the attention function.

%% file: figs/cross_attn.tex
\begin{figure}
    \centering
    \includegraphics[width=0.85\linewidth]{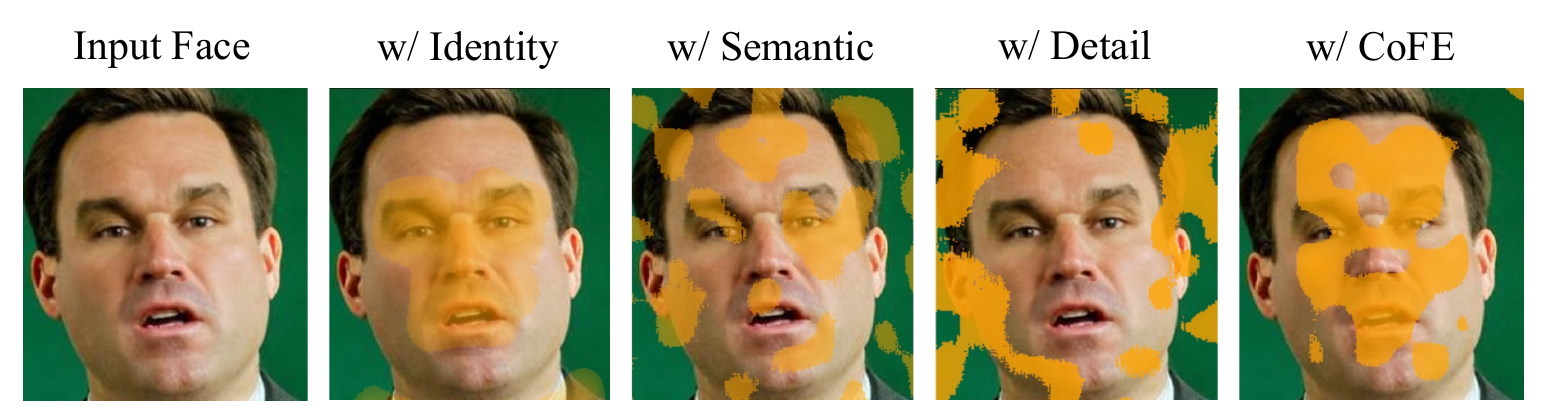}
    \vspace{-2mm}
    \caption{Cross-attention visualizations of different experts. Warmer colors denote higher attention strength. \method~integrates complementary attention from Identity Expert (ArcFace), Semantic Expert (CLIP), and Detail Expert (DINOv2) to jointly enhance facial detail fidelity and identity consistency.}
    \label{fig:cross_attn}
    \vspace{-4mm}
\end{figure}
% The input images are from OpenHumanVid~\cite{li2025openhumanvid}.

%% file: tables/dataset_comp.tex
\begin{table*}[t]
\centering
\caption{Comparison with existing video datasets. Our \ourdata-180K dataset uniquely provides \textbf{identity-consistent and clustered} human videos with skeleton and face pose annotations, supporting high-fidelity identity preservation in video generation.}
\label{tab:datasets}
\begin{tabular}{lccccccc}
\toprule
Dataset name & Domain & \# Videos & Annotations & Resolution & ID-Consistent? & ID-Clustered? \\
\midrule
Koala-36M~\cite{wang2025koala} & Open & 36M  & - & 720P & \usym{2717} & \usym{2717} \\
OpenHumanVid~\cite{li2025openhumanvid} & Human & 13.2M & Skeleton pose & 720P & \usym{2717} & \usym{2717}  \\
\ourdata-180K(Ours) & Human & 180K & Skeleton pose, face pose & 720P, 1080P & \usym{2713} & \usym{2713} \\
\bottomrule
\end{tabular}
\label{tab:dataset_comp}
\end{table*}

%% file: secs/4_dataset.tex
\section{\ourdata~Benchmark}
% 本章节参考openhumanvid进行组织 
\input{figs/data_count}

% We adopt OpenHumanVid~\cite{li2025openhumanvid} and additional in-house data to curate \textbf{\ourdata-180K}, a large-scale dataset designed for identity-preserving video generation under challenging facial pose variations. 

% \ourdata-180K leverages the high-quality, human-centric video foundation of OpenHumanVid and enriches it with curated in-house content. The resulting clips are 2–6 seconds long, captured predominantly at 720P or 1080P resolution to ensure high visual fidelity, and exhibit rich diversity in human appearance, pose, and motion. Each clip is accompanied by a detailed caption (typically 45–75 words), refined to explicitly mark subject speech and thereby reduce lip-speech ambiguity in cinematic content. Word cloud analysis of the captions reveals dominant human-centric terms such as ``man'', ``woman'', ``dressed'', and ``wear'', underscoring the dataset’s explicit focus on diverse human attributes.
\subsection{Overview}

To advance the study and evaluation of identity-preserving video generation under large face poses, we introduce \textbf{\ourdata-180K}, a benchmark derived from OpenHumanVid~\cite{li2025openhumanvid} and supplemented with self-collected high-quality videos. As illustrated in Fig. \ref{fig:datacounts}, the dataset comprises 180K short clips (2$\sim$6 seconds, 720P$\sim$1080P) featuring diverse identities, poses, and motions. Each clip is paired with a refined caption (45$\sim$75 words) that explicitly annotates spoken content and appearance attributes, supporting both semantic grounding and lip-sync disambiguation. This design ensures \ourdata~serves not only as a scalable training resource but, more critically, as a challenging benchmark for identity-preserving video generation. 

A key design goal is to ensure both consistent identity representation and rich dynamic pose variation. \wyj{To this end, our processing pipeline employs objective facial quality filtering that naturally yields videos with high aesthetic appeal, strong imaging quality, and stable subject identity(Fig.~\ref{fig:dataquality}).} We quantify pose variation by estimating 3D face angles (pitch, yaw, roll) per frame and computing the maximum angular range within each clip. As Fig.~\ref{subfig:angle} demonstrates, \ourdata~exhibits significantly broader pose coverage than existing datasets, especially in extreme yaw and pitch, making it uniquely suited for identity preservation under challenging face motions.

\wyj{Existing benchmarks such as VIP-Test~\cite{vip-200k} focus on full-body motion but lack sufficient variation in large face poses, limiting their ability to evaluate identity preservation under challenging pose conditions. To address this gap, we introduce \textbf{\ourdata-Bench}, a standardized evaluation protocol derived from \ourdata-180K, comprising fifty held-out identities (disjoint from training) with four diverse prompts per identity, all featuring large face poses that span extreme yaw and pitch ranges. For generalization assessment, we additionally construct a VIP-Test subset using SDXL-generated~\cite{podell2023sdxl} reference images (fifty identities, four prompts each). Together, these two test sets enable the first holistic evaluation of image-to-video models across both standard full-body animation and the challenging regime of identity preservation under large face poses.}

\input{figs/data_quality}

\input{figs/data_processing}

\subsection{Data Processing Pipeline}
Our overall data processing pipeline is illustrated in Fig.~\ref{fig:datapipeline}. To collect videos with substantial face pose variations, we apply a multi-stage procedure.
The first step, \textbf{Face Constraints Filtering}, retains clips exhibiting large face pose variations (\textit{e.g.}, extreme pitch, yaw or roll rotations). The second step, \textbf{Identity Consistency Analysis}, ensures facial identity remains coherent during pronounced face motion by discarding clips with inconsistent identity across frames. The third step, \textbf{Speech Disambiguation}, aligns lip motion or speech activity with caption by annotating whether the on-screen subject is speaking.
Together, these steps ensure high-quality data for downstream identity-preserving tasks.

\input{tables/main_comp_easy}
\input{tables/main_comp_hard}

\noindent \textbf{Face Constraints Filtering.} 
To support identity preservation under diverse face poses, especially large face motions, we curate high-quality samples for \ourdata~via multi-dimensional face constraints filtering, enforcing strict criteria on face continuity, visibility, and pose diversity. For efficiency, we sample one frame every three and detect face bounding boxes and poses using ArcFace~\cite{deng2020retinaface}. The core constraints are:
\textbf{i) Face count.} Focused on single person generation, we filter videos where any selected frame contains no faces or multiple faces. This ensures a single, continuous face throughout the video clip, foundational for consistent identity learning. 
\textbf{ii) Face proportion.} Video clips with low face bounding box proportion (\ie, less than 10\%) are filtered out to emphasize the preservation of facial details.
\textbf{iii) Face pose diversity.} To ensure preservation of identity across all face poses, we retain videos with diverse poses. We analyze the face poses of selected frames for each video clip, selecting videos with large face pose variations.
% \end{itemize}

\noindent \textbf{Identity Consistency Analysis}
Identity consistency is essential for human-centric videos, reflecting the stability of facial representations across frames. To ensure high-quality data curation, we analyze identity coherence using ArcFace features~\cite{deng2019arcface}.
\textbf{i) Identity Similarity.} We compute pairwise cosine similarities between frame embeddings $\{\mathbf{d}_i\}_{i=1}^N$:
\begin{equation}
\text{sim}(\mathbf{d}_i, \mathbf{d}_j) = \frac{\mathbf{d}_i \cdot \mathbf{d}_j}{\|\mathbf{d}_i\| \|\mathbf{d}_j\|} \quad (i \neq j),
\end{equation}
and use the average off-diagonal similarity $\overline{s}$ as a measure of intra-clip identity coherence. Clips with $\overline{s} \leq 0.6$ are discarded.
\textbf{ii) Identity Clustering.} To ensure disjoint identity splits for fair train/test evaluation, we cluster videos by identity using FAISS~\cite{douze2024faiss}. An IVF-PQ index~\cite{xiao2022distill} enables efficient similarity search in the ArcFace embedding space via product quantization. A two-stage hierarchical strategy is applied: a high threshold first forms tight identity clusters from strongly correlated samples; a lower threshold then assigns ambiguous samples to recover weakly correlated but same-identity instances.

% Identity consistency is critical for human-centric videos, directly reflecting facial representation quality. To enhance data quality during cleaning, we introduce an identity consistency analysis mechanism, using ArcFace features~\cite{deng2019arcface,deng2020retinaface}. 

% \textbf{i) Identity Similarity.} We construct a pairwise similarity matrix based on cosine similarity:
% \begin{equation}
% \text{sim}(\mathbf{d}_i, \mathbf{d}_j) = \frac{\mathbf{d}_i \cdot \mathbf{d}_j}{|\mathbf{d}_i| |\mathbf{d}_j|} \quad (i \neq j),
% \end{equation}
% where $\{\mathbf{d}_i\}^N_{i=1}$ represent the ArcFace features for $N$ frame. We derive the average similarity \(\overline{s}\) as the mean of all off-diagonal elements, which serves as an indicator of the overall identity coherence for one video clip. We then retain only videos where \(\overline{s}\) exceeds a threshold of 0.6, ensuring the final dataset contains samples with stable and coherent identity representations to enhance the reliability of downstream tasks. 
% \textbf{ii) Identity Clustering.} To ensure disjoint identity splits for fair train/test evaluation, we cluster videos by identity using FAISS~\cite{douze2024faiss}. We build an IVF-PQ~\cite{xiao2022distill} index to enable efficient similarity search in the high-dimensional ArcFace embedding space via product quantization. A two-stage hierarchical strategy is applied: a high threshold first forms tight identity clusters from strongly correlated samples; a lower threshold then assigns remaining samples to recover weakly correlated but same-identity instances. 

\noindent \textbf{Speech Disambiguation.}
\ourdata~is mainly curated from OpenHumanVid~\cite{li2025openhumanvid}, where most videos show people speaking with continuous lip movements but captions rarely indicate actual speech or lip motion. Without explicit speech semantics, identity-preserving generators often wrongly link lip motion to text, producing videos with incessant lip movement even when no speech motion is contained in the input caption. To align lip motion with caption, we disentangle speaking status in two steps: whisper~\cite{radford2023robust} detects audible speech and its alignment with the primary subject’s lip movements; if the caption lacks explicit description of speaking, an LLM~\cite{bai2023qwen,touvron2023llama} appends a context-aware sentence (\textit{e.g.}, ``The man is talking.'') to signal active speech. This ensures genuine speaking behavior is reflected in text-video alignment, enabling controllable and identity-consistent generation.

% \wyj{Since \ourdata~derives from film clips, many videos show pronounced lip movements without corresponding speech from the on-screen subject. To prevent identity generators from conflating non-verbal lip motion with textual semantics, we disentangle speaking status in two steps. First, we use whisper~\cite{radford2023robust} to detect whether speech occurs in the audio and whether it temporally aligns with the primary subject’s lip movements. Second, we inspect the original caption: if it lacks explicit indication of speaking, we employ Qwen~\cite{bai2023qwen} to append a natural, context-aware sentence (\textit{e.g.,}, ``The man is talking.'') to explicitly signal active speech. This ensures text-video alignment reflects true speaking behavior, reducing ambiguity in identity-preserving training.}

%% file: figs/data_count.tex
\begin{figure}[t]
  \centering
  \begin{subfigure}[b]{0.47\linewidth}
    \centering
    \includegraphics[width=\linewidth]{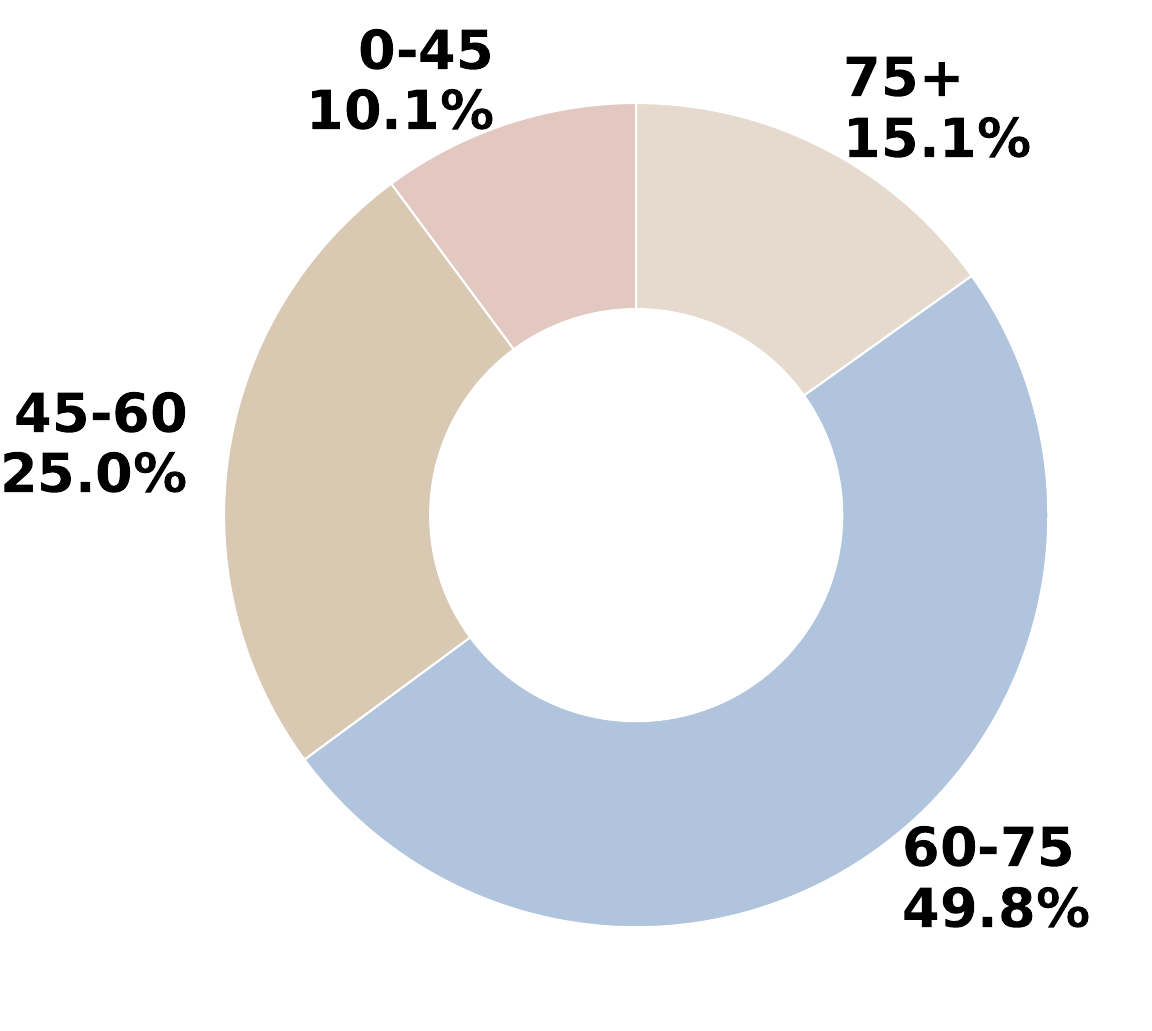}
    \caption{Caption Word Count}
    \label{subfig:prompt_word_count}
  \end{subfigure}
  \hfill
  \begin{subfigure}[b]{0.43\linewidth}
    \centering
    \includegraphics[width=\linewidth]{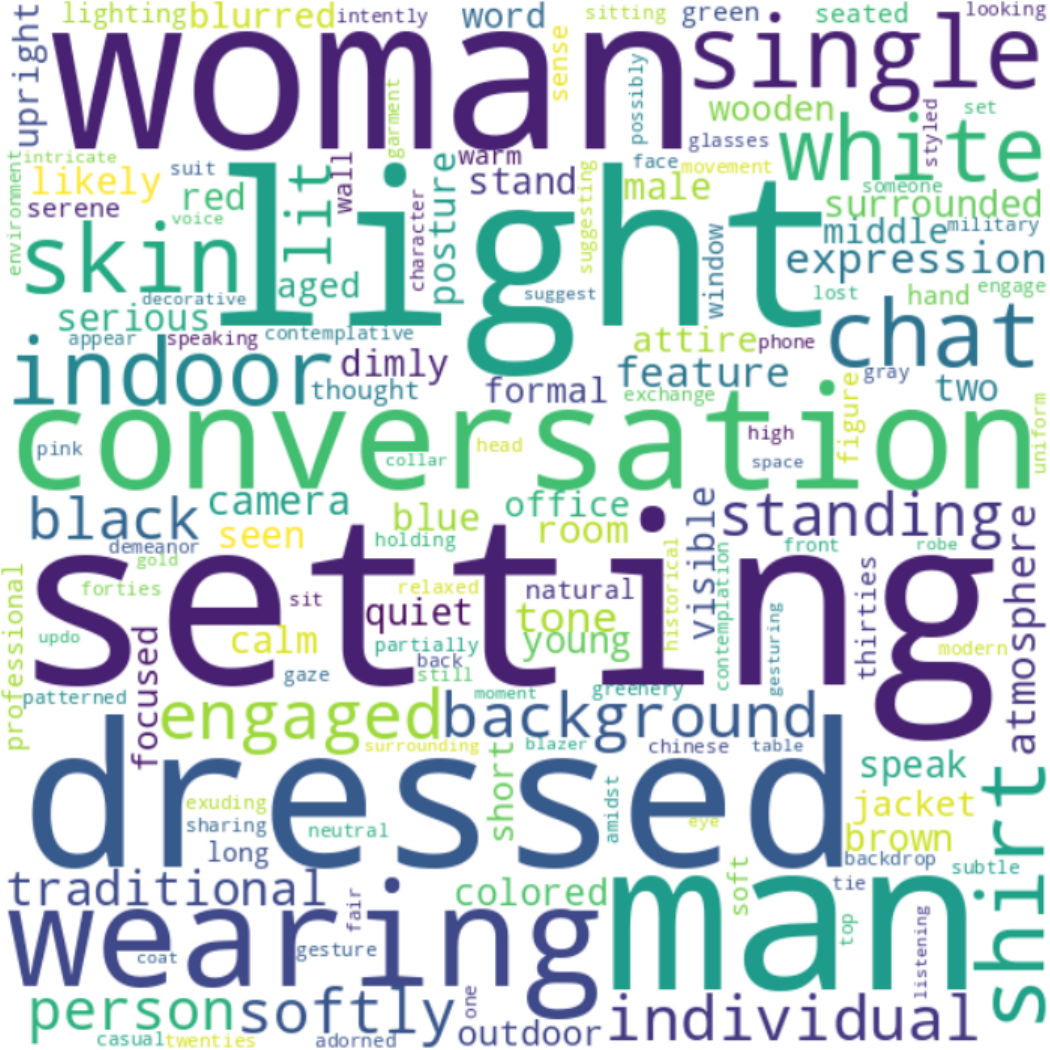}
    \caption{Caption Word Cloud}
    \label{subfig:prompt_wordcloud}
  \end{subfigure}

  \begin{subfigure}[b]{0.49\linewidth}
    \centering
    \includegraphics[width=\linewidth]{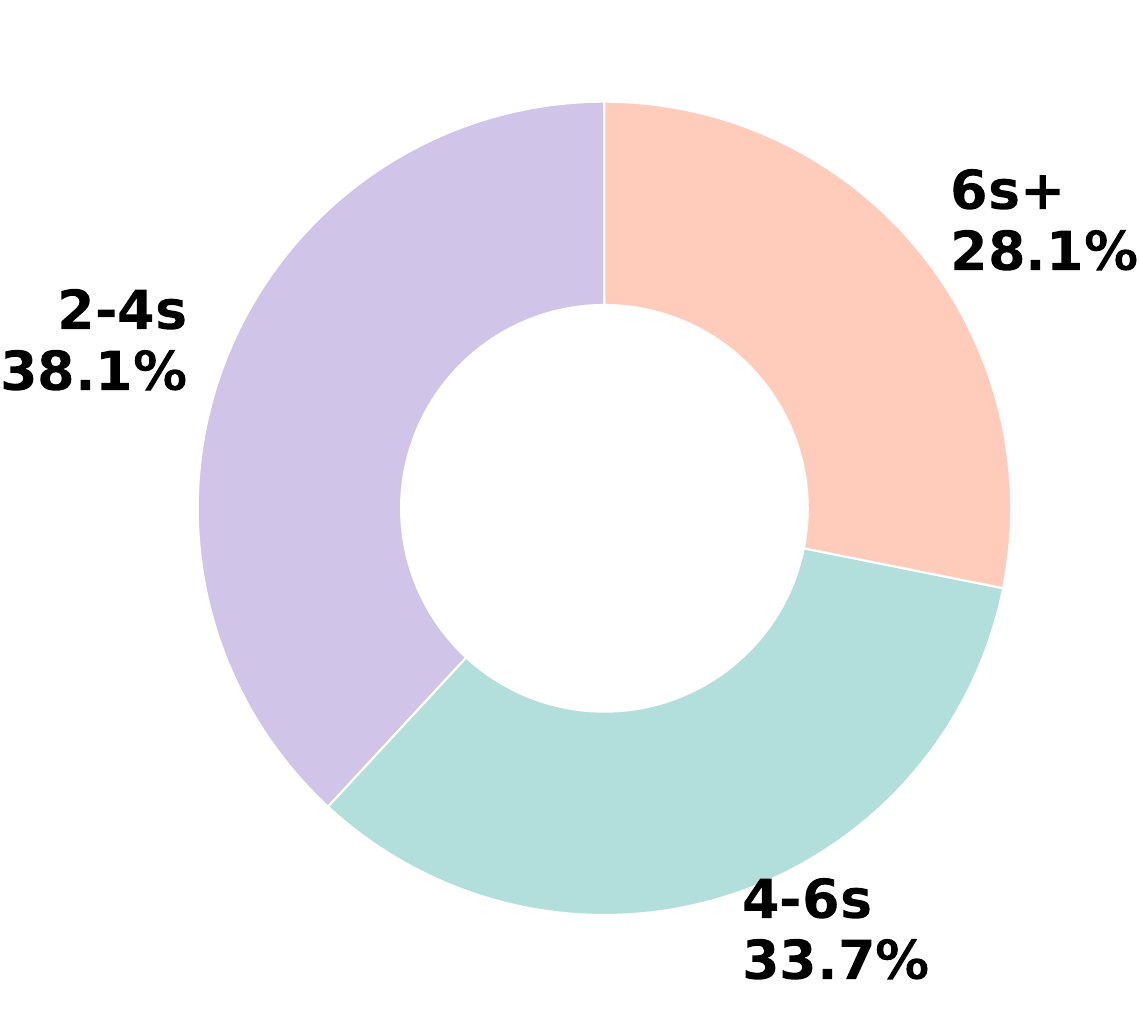}
    \caption{Video Duration}
    \label{subfig:video_duration}
  \end{subfigure}
  \hfill
  \begin{subfigure}[b]{0.43\linewidth}
    \centering
    \includegraphics[width=\linewidth]{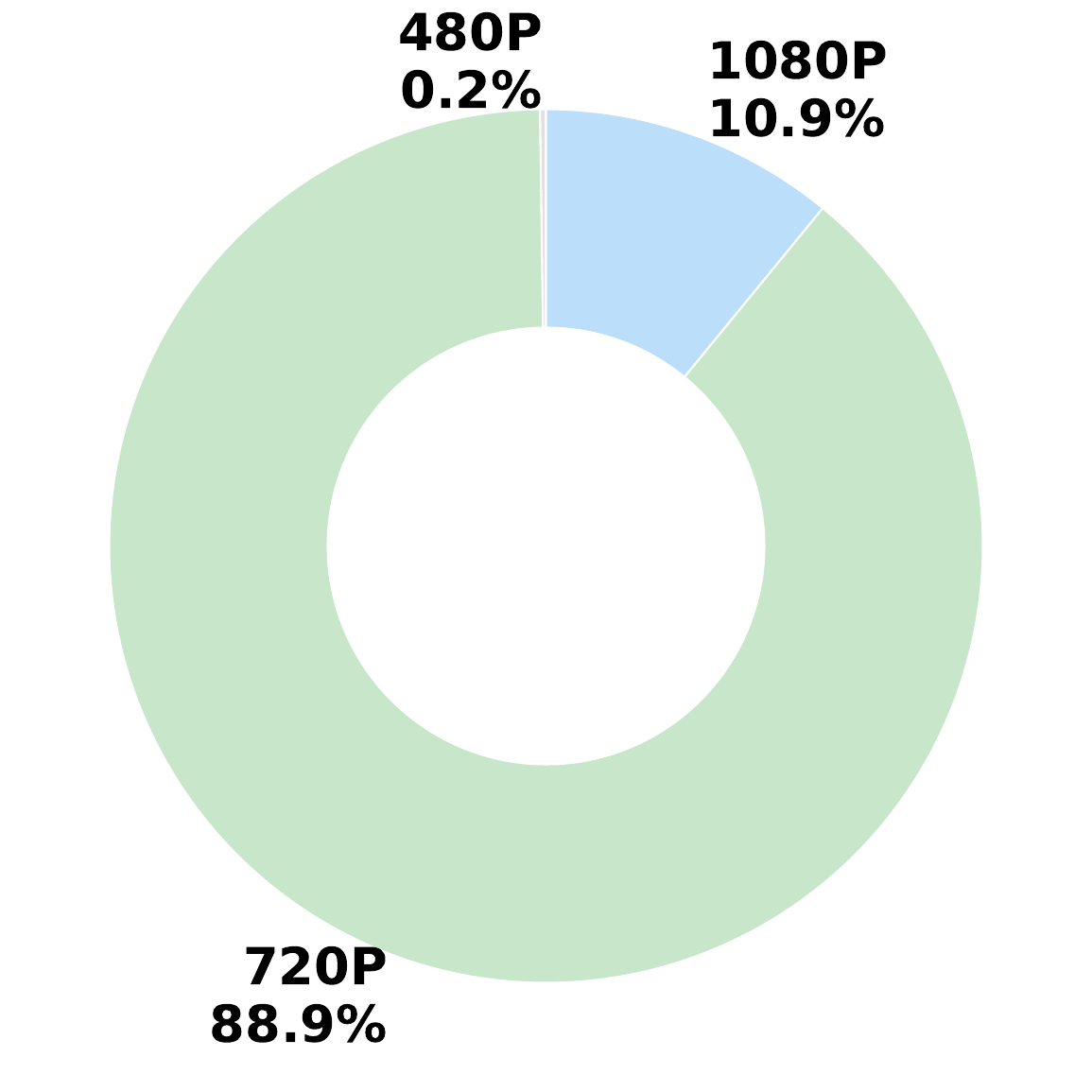}
    \caption{Video Resolution}
    \label{subfig:video_resolution}
  \end{subfigure}

  \caption{The statistical analysis of the \ourdata-180K dataset.}
  \label{fig:datacounts}
\end{figure}

%% file: figs/data_quality.tex
\begin{figure}[t]
 \begin{subfigure}[b]{0.98\linewidth}
    \centering
    \includegraphics[width=\linewidth]{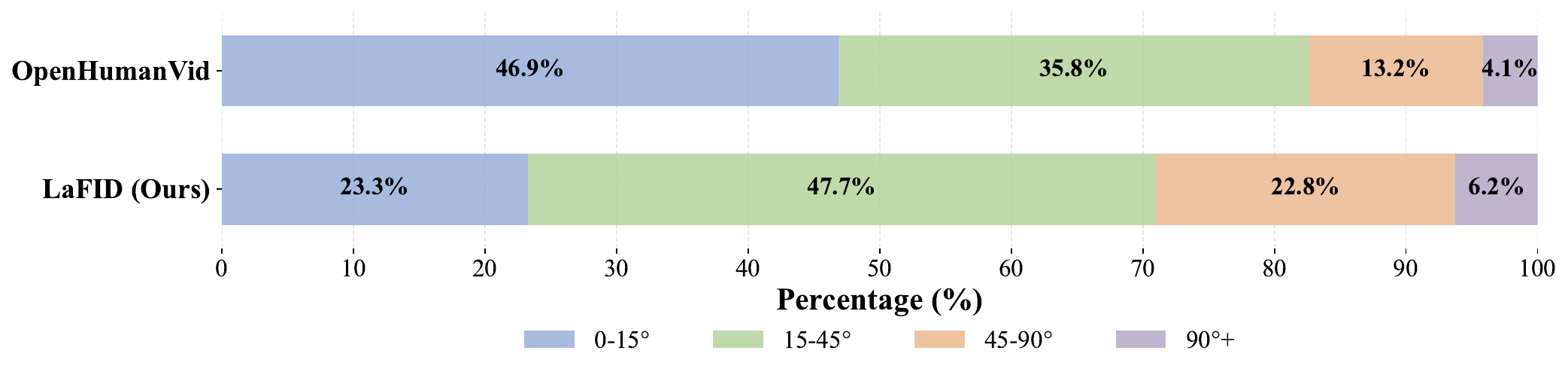}
    \caption{Face Pose Variations}
    \label{subfig:angle}
  \end{subfigure}
  \centering
  \begin{subfigure}[b]{0.48\linewidth}
    \centering
    \includegraphics[width=\linewidth]{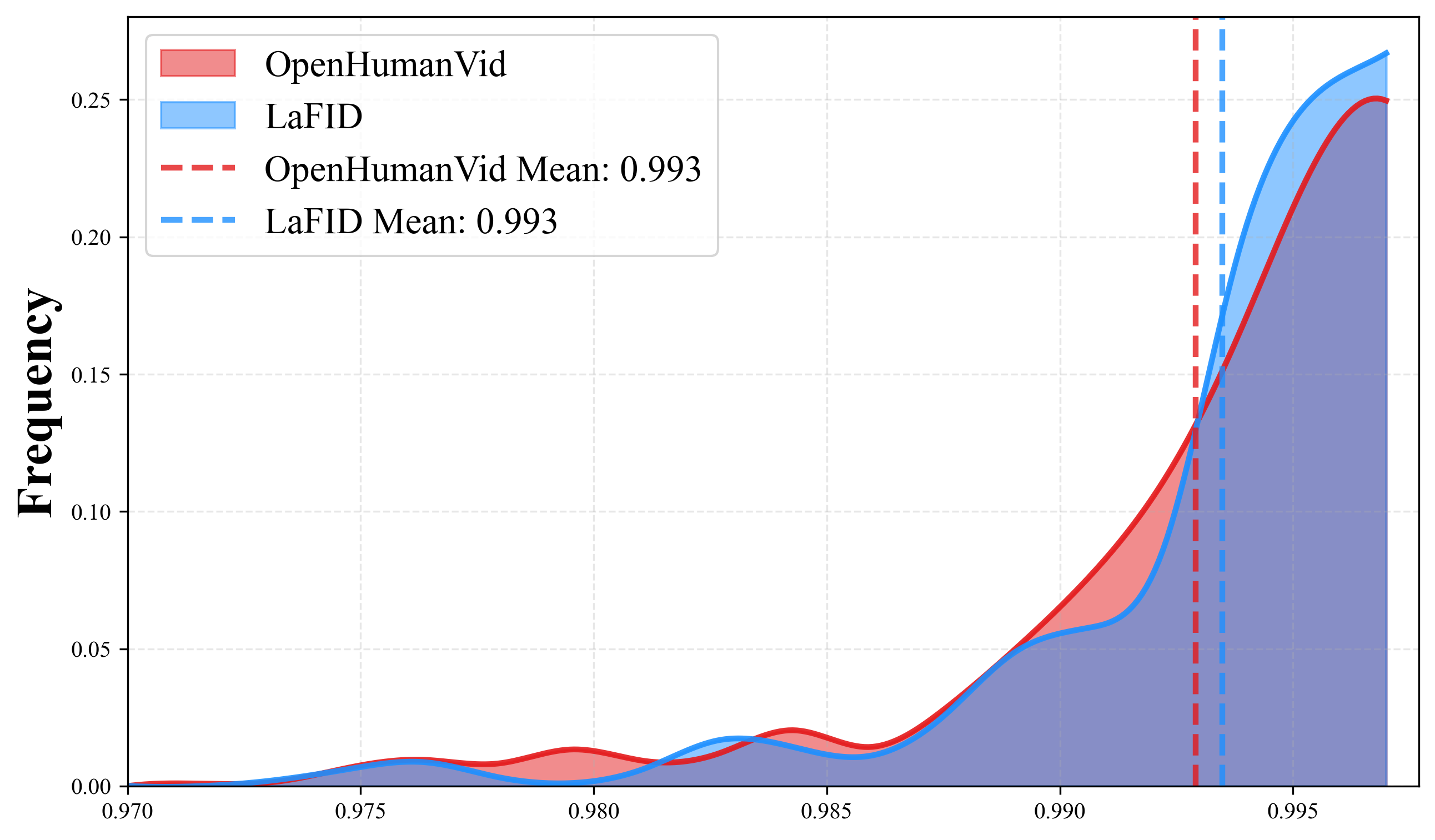}
    \caption{Subject Consistency}
    \label{subfig:sub}
  \end{subfigure}
  \hfill
  \begin{subfigure}[b]{0.48\linewidth}
    \centering
    \includegraphics[width=\linewidth]{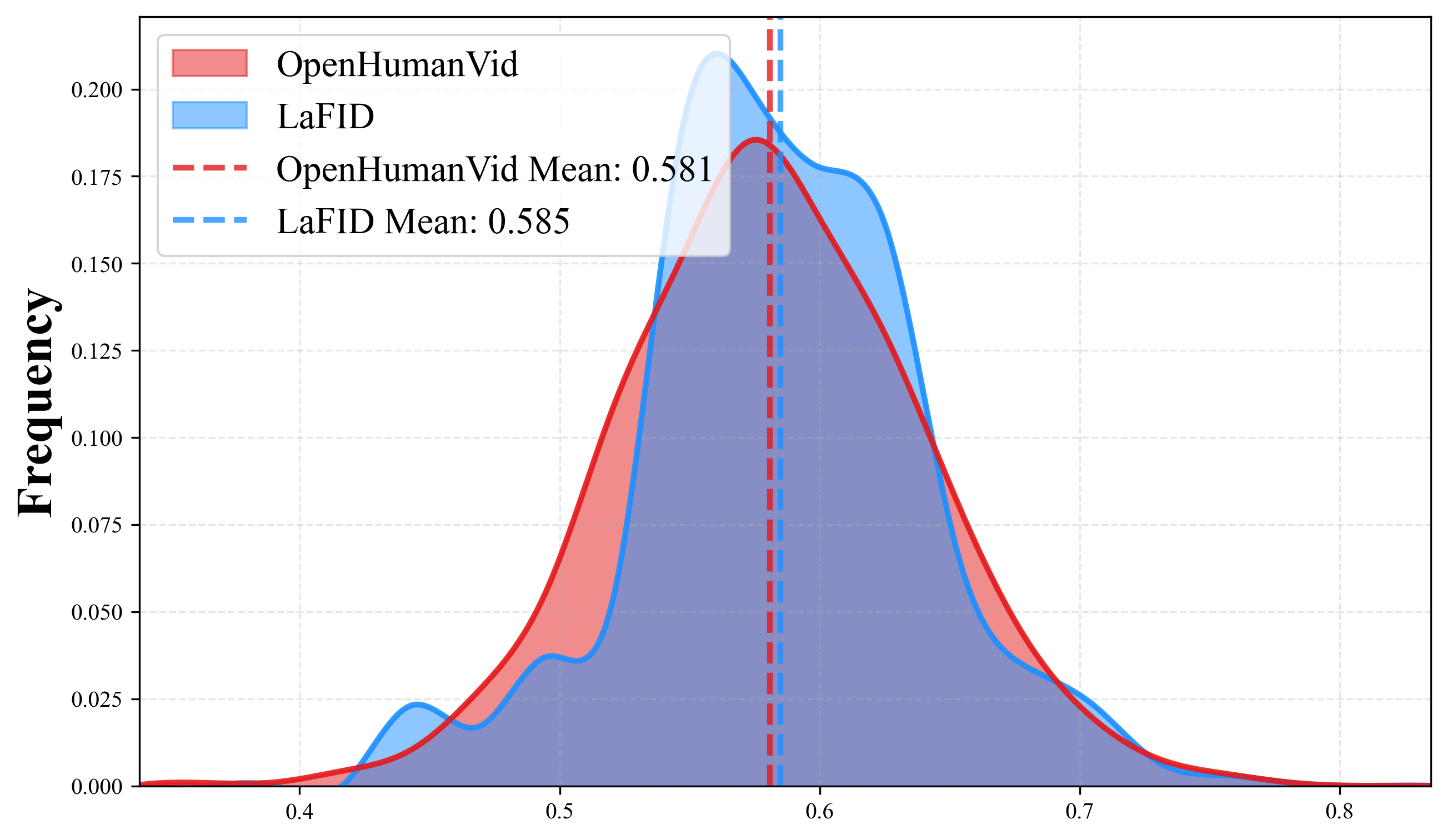}
    \caption{Aesthetic Score}
    \label{subfig:aes}
  \end{subfigure}

  \begin{subfigure}[b]{0.48\linewidth}
    \centering
    \includegraphics[width=\linewidth]{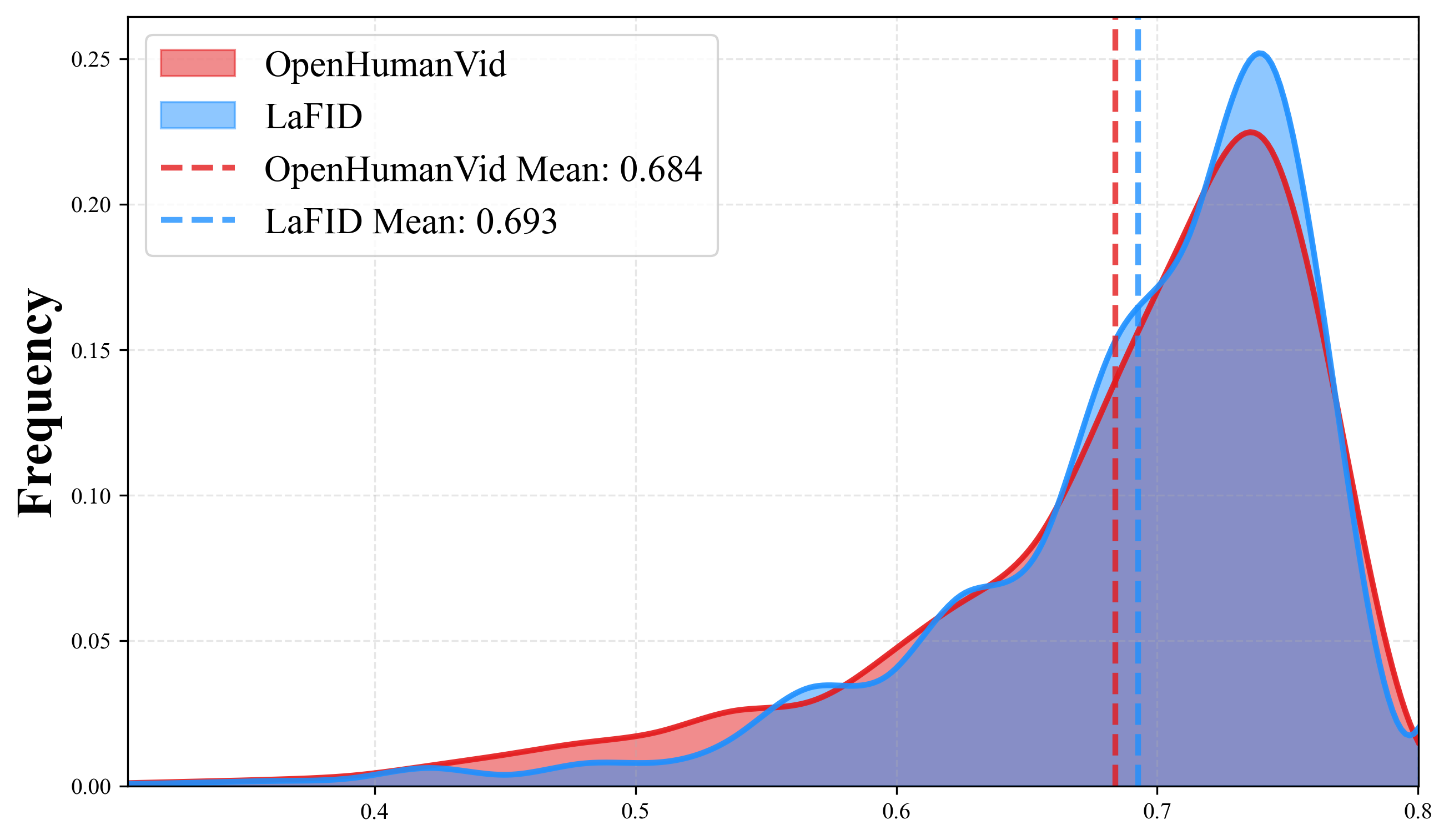}
    \caption{Imaging Quality}
    \label{subfig:imaging}
  \end{subfigure}
  \hfill
  \begin{subfigure}[b]{0.48\linewidth}
    \centering
    \includegraphics[width=\linewidth]{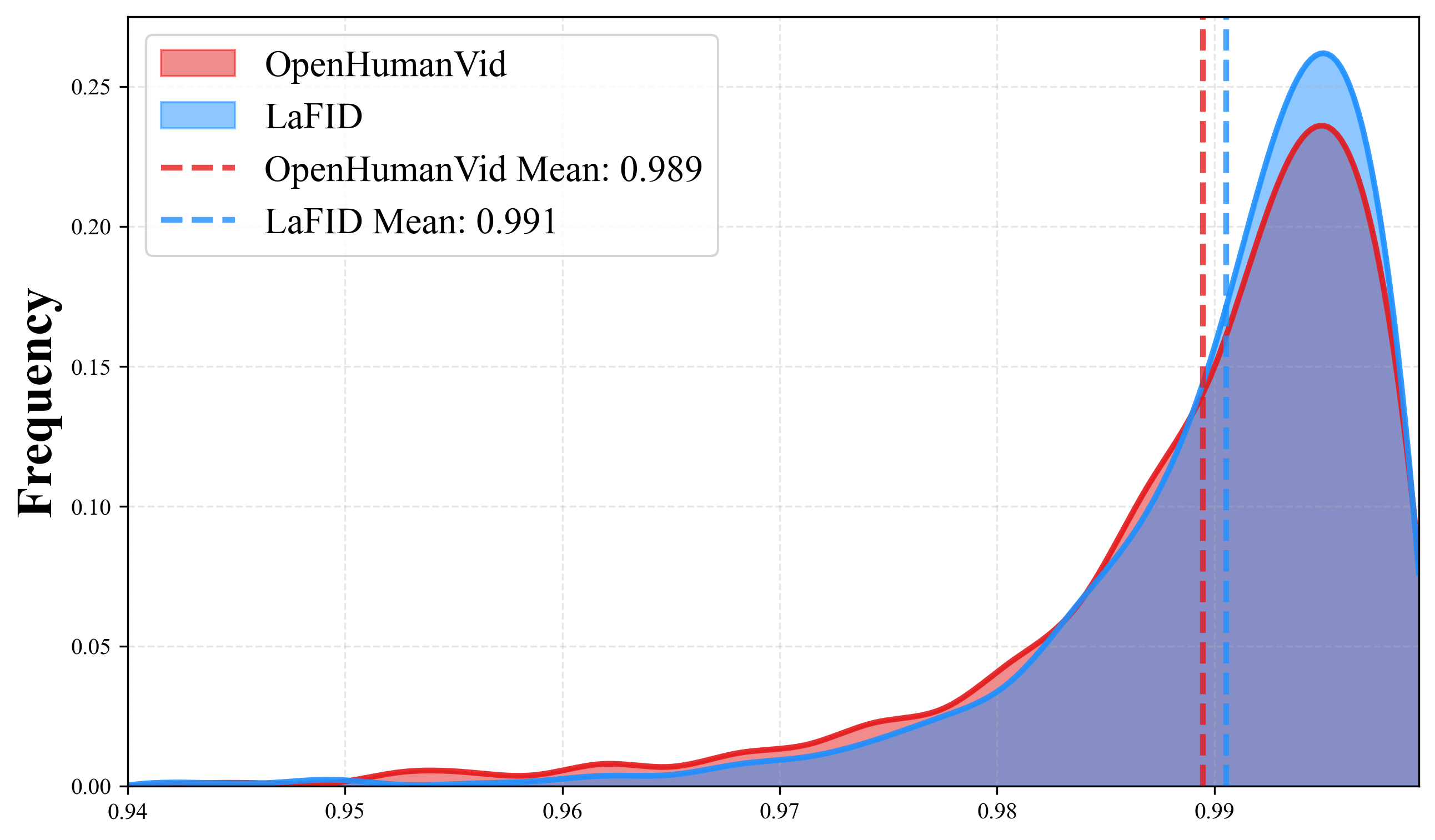}
    \caption{Time Flickering}
    \label{subfig:time}
  \end{subfigure}
  \caption{Comparison of video quality metrics between \ourdata-180K and OpenHumanVid. \ourdata-180K exhibits better quality.}
  \label{fig:dataquality}
  % \vspace{-4mm}
\end{figure}

%% file: figs/data_processing.tex
\begin{figure}
    \centering
    \includegraphics[width=0.9\linewidth]{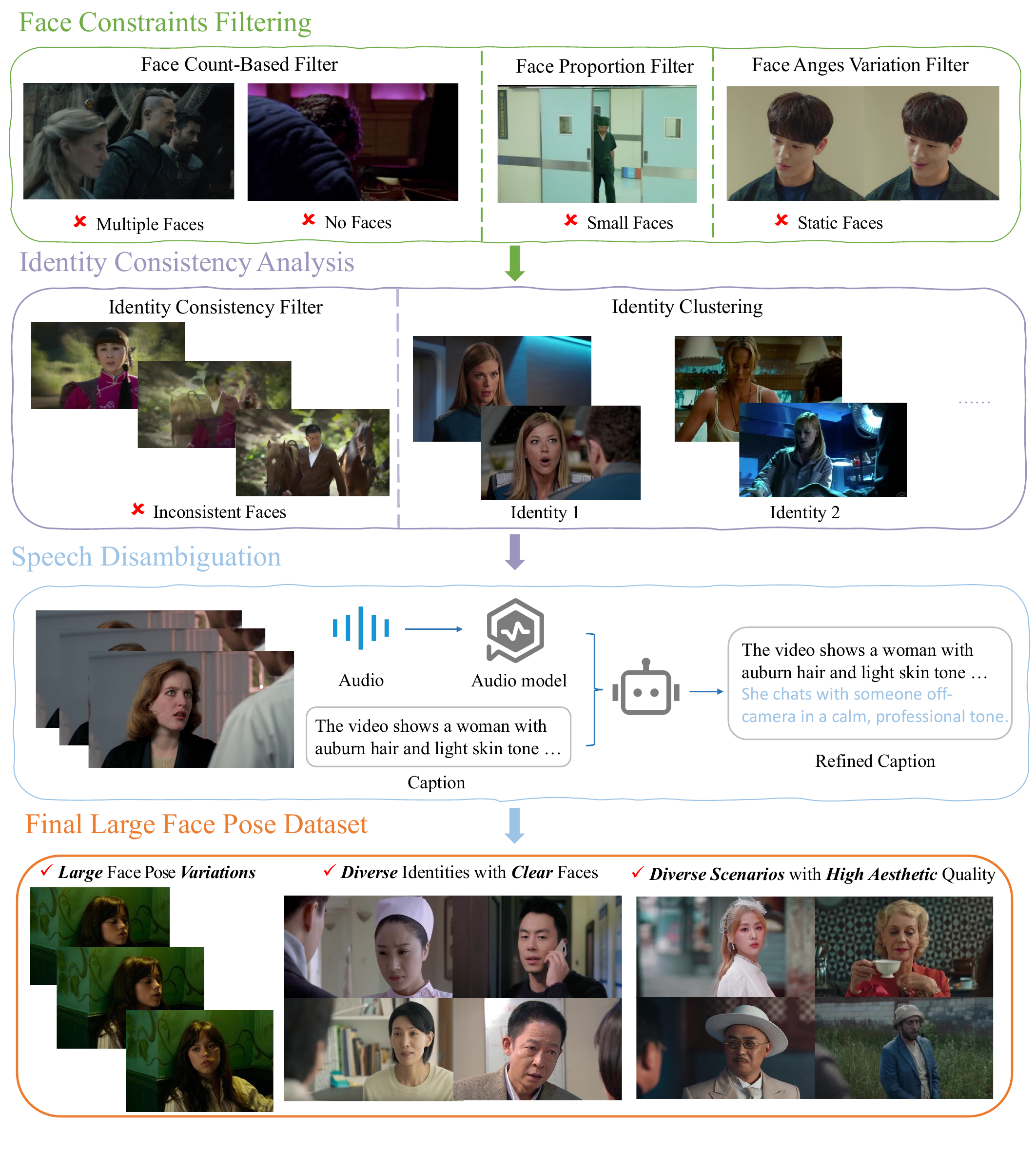}
    \caption{The overview of the data processing pipeline for constructing dataset with large face poses.}
    \label{fig:datapipeline}
    \vspace{-4mm}
\end{figure}
% The input images are from OpenHumanVid~\cite{li2025openhumanvid}.

%% file: tables/main_comp_easy.tex
\begin{table*}[tbp]
\renewcommand{\tabcolsep}{4pt}
\centering
\caption{Comparison with SOTA methods on \ourdata-Bench benchmark dataset. The best results are highlighted in \textbf{Bold} ($\uparrow$/$\downarrow$ represents the higher/lower the better). CoFE (Ours) has achieved the best performance across most metrics, demonstrating its superiority. }
\begin{tabular}{lccccccccc}
\hline
\hline
\multirow{2}{*}{Method} & \multicolumn{2}{c}{Identity Preservation} & Text Relevance & \multicolumn{5}{c}{Video Quality} \\ \cline{2-3} \cline{5-9}
& FaceSim-Arc↑ & FaceSim-Cur↑ & CLIP-Score↑ & FID↓ & AQ↑ & IQ↑ & TF↑ & MS↑ \\ \hline
% ConsisID~\cite{yuan2025consisid} & 0.567 & 0.573 & \textbf{31.184} & 123.967 & 0.560 & 0.670 & 0.979 & 0.990 \\
CogVideoX-5B~\cite{yang2024cogvideox} & 0.629 & 0.636 & 29.331 & 104.214 & 0.499 & 0.598 & 0.976 & 0.986 \\
LTX-2B~\cite{hacohen2024ltx} & 0.438 & 0.440 & 28.800 & 151.773 & 0.513 & 0.581 & 0.980 & 0.992 \\
ConsisID~\cite{yuan2025consisid} & 0.567 & 0.573 & \textbf{31.184} & 123.967 & 0.560 & 0.670 & 0.979 & 0.990 \\
Wan2.1-I2V-1.3B~\cite{wan2025wan} & 0.654 & 0.656 & 29.200 & 72.322 & 0.553 & 0.660 & 0.989 & 0.993  \\ 
Wan2.1 w/ \textbf{\method} & 0.770 & 0.775 & 29.125 & \textbf{63.799} & 0.570 & 0.673 & 0.990 & 0.994 \\
Wan2.2-TI2V-5B~\cite{wan2025wan} & 0.776 & 0.786 & 28.761 & 76.728 & 0.588 & 0.666 & 0.995 & \textbf{0.996} \\
Wan2.2 w/ \textbf{\method} & \textbf{0.814} & \textbf{0.827} & 28.729 & 70.727 & \textbf{0.591} & \textbf{0.669} & \textbf{0.996} & \textbf{0.996}  \\
\hline
\hline
\end{tabular}
\label{tab:main_comp_easy}
\end{table*}

%% file: tables/main_comp_hard.tex
\begin{table*}[tbp]
\renewcommand{\tabcolsep}{4pt}
\centering
\caption{Comparison with SOTA methods on the VIP-Test benchmark. Our \method~markedly improves these metrics, achieving the best identity consistency while maintaining strong text and video quality.}
\begin{tabular}{lccccccccc}
\hline
\hline
\multirow{2}{*}{Method} & \multicolumn{2}{c}{Identity Preservation} & Text Relevance & \multicolumn{5}{c}{Video Quality} \\ \cline{2-3} \cline{5-9}
& FaceSim-Arc↑ & FaceSim-Cur↑ & CLIP-Score↑ & FID↓ & AQ↑ & IQ↑ & TF↑ & MS↑ \\ \hline
CogVideoX-5B~\cite{yang2024cogvideox} & 0.462 & 0.474 & 28.947 & 163.001 & 0.512 & 0.585 & 0.926 & 0.948 \\
LTX-2B~\cite{hacohen2024ltx} & 0.301 & 0.323 & 28.403 & 203.516 & 0.510 & 0.586 & 0.944 & 0.975 \\
ConsisID~\cite{yuan2025consisid} & 0.269 & 0.289 & 29.601 & 206.924 & 0.566 & 0.638 & 0.946 & 0.976 \\
Wan2.1-I2V-1.3B~\cite{wan2025wan} & 0.471 & 0.480 & 29.825 & 129.116 & 0.554 & 0.659 & 0.940 & 0.967 \\ 
Wan2.1 w/ \textbf{\method} & 0.502 & 0.510 & \textbf{30.167} & 126.676 & 0.558 & \textbf{0.679} & 0.947 & 0.971 \\
Wan2.2-TI2V-5B~\cite{wan2025wan} & 0.541 & 0.552 & 29.819 & 125.893 & 0.574 & 0.667 & 0.934 & 0.964 \\
Wan2.2 w/ \textbf{\method} &  \textbf{0.605} & \textbf{0.615} & 29.834 & \textbf{113.850} & \textbf{0.583} & 0.668 & \textbf{0.948} & \textbf{0.979} \\
\hline
\hline
\end{tabular}
\vspace{-4mm}
\label{tab:main_comp_hard}
\end{table*}

%% file: secs/5_experiment.tex
\section{Experiments}

\input{figs/qualitative_analysis}

\subsection{Implementation Details}
\subsubsection{Training strategies} 
We integrate the proposed \method~into the Wan2.1-I2V-1.3B and Wan2.2-TI2V-5B models~\cite{wan2025wan}. 
To balance performance and efficiency, we inject CoFE into each even-numbered DiT block, in line with VACE~\cite{jiang2025vace}.
This design enhances model capabilities while incurring a modest parameter increase of merely 18.5\%. We employ aspect ratio bucketing for multi-resolution and multi-frame training, with a minimum video resolution of 480p and a frame count ranging from 49 to 129. To address the ``copy-paste" shortcut, we incorporate face features from non-training frames into the input for video clips with additional frames. We use the Adam optimizer with a learning rate of 1e-5 while keeping other modules frozen, and train on the \ourdata-180K dataset for 4,000 iterations with a batch size of 256.

\subsubsection{Evaluation Protocols}
We evaluate our method on \ourdata-Bench and VIP-Test~\cite{vip-200k}, adopting the evaluation protocol of ConsisID~\cite{yuan2025consisid} for metric selection. Specifically, we assess identity preservation using FaceSim-Arc~\cite{deng2019arcface} and FaceSim-Cur~\cite{huang2020curricularface}, text relevance via CLIP-Score~\cite{hessel2021clipscore}, facial fidelity through FID, and overall video quality with VBench~\cite{huang2024vbench}. The latter includes four key sub-metrics: Aesthetic Quality (AQ), Imaging Quality (IQ), Temporal Flickering (TF), and Motion Smoothness (MS). 

\subsection{Benchmark Results}
We compare our method with open-source state-of-the-art approaches~\cite{yang2024cogvideox,hacohen2024ltx,yuan2025consisid,wan2025wan} on \ourdata-Bench and VIP-Test~\cite{vip-200k} to demonstrate the superiority of our \method~module in preserving identity under large face poses.

\subsubsection{Quantitative Evaluation}
\wyj{As shown in Tables~\ref{tab:main_comp_easy} and~\ref{tab:main_comp_hard}, our \method~delivers substantial improvements on both Wan2.1 and Wan2.2, with larger gains for Wan2.1 on \ourdata-Bench and for Wan2.2 on VIP-Test. This consistent enhancement across model generations and generation scenarios reveals a key insight: identity preservation relies not on global video modeling alone but on the explicit injection of pose-robust, disentangled facial representations. Notably, our approach remains effective even in full-body settings where faces are small or temporally unstable, demonstrating that high-fidelity identity control can be achieved through localized facial priors decoupled from full-body motion synthesis.}

\wyj{On general video quality metrics (AQ, IQ, TF, and MS), \method~outperforms all baselines, including Wan2.2, yielding higher visual appeal, fewer artifacts, and smoother motion. Existing methods exhibit distinct limitations: LTX-2B suffers from identity drift; ConsisID achieves high CLIP-Score but often generates blurry or distorted faces due to its narrow focus on facial alignment; and CogVideoX-5B delivers only moderate performance despite its large scale. Even the improved Wan2.2 falls short of \method~in both identity fidelity and overall quality. These results demonstrate that explicitly modeling disentangled facial attributes such as identity, semantics, and detail as structured priors enables more robust and coherent video generation under complex dynamics.}

\input{tables/abla_1}
% \inputtable{abla_2}
\input{figs/visual_ablation}

\subsubsection{Qualitative Evaluation} 
Fig. \ref{fig:main_exp} presents qualitative comparison between our method and other SOTA methods. In scenarios with significant face movements, our model exhibits more stable generation and superior identity preservation: multi-source feature injection enhances facial details, ensuring consistent identity throughout the video clip. 
Specifically, ConsisID poorly renders front-facing details when given only a profile as input. CogVideoX and Wan2.2 fail to preserve identity during large face movements, causing obvious distortions and identity shifting. LTX-Video generates unstable, often blurry frames with lost identity details. Supplementary videos further validate our approach’s robustness. 

\subsection{Ablation Analysis}
\noindent \wyj{\textbf{Feature Fusion Strategy Comparison.} To validate the advantages of our adaptive face feature fusion, we compare \method~with two common baselines: single-expert injection using the identity, semantic, or detail expert alone, and naive concatenation of all three experts.  
We evaluate these strategies on \ourdata-Bench using Wan2.1-I2V-1.3B as the base model. As shown in Table~\ref{tab:abla_1} and Fig.~\ref{fig:abla}, single-expert variants improve over the baseline but exhibit distinct weaknesses: the identity expert lacks fine details such as eye shape, the semantic expert shows inconsistent skin tones, and the detail expert yields blurry contours under large poses. Naive concatenation aggregates cues but treats them uniformly, introducing redundancy and misalignment that harm identity consistency. In contrast, \method~achieves the highest FaceSim-Arc and FaceSim-Cur scores and the lowest FID, confirming that adaptive fusion of complementary facial representations is key to robust identity preservation.}

\input{tables/abla_datasets}

\noindent \textbf{Ablation of Training Data Processing.} In our data processing pipeline, we introduce Face Constraints Filtering and Speech Disambiguation to enhance identity stability and semantic alignment. As shown in Table~\ref{tab:abla_datasets}, models trained on our filtered data consistently outperform those using the raw OpenHumanVid subset, validating the effectiveness of our proposed data processing pipeline. Furthermore, incorporating Speech Disambiguation yields additional gains across all metrics, demonstrating that explicitly aligning captions with speaking status helps mitigate spurious lip-motion cues and improves identity preservation. More qualitative results are provided in our supplementary material.

%% file: figs/qualitative_analysis.tex
\begin{figure*}[t]
    \centering
    \includegraphics[width=0.9\linewidth]{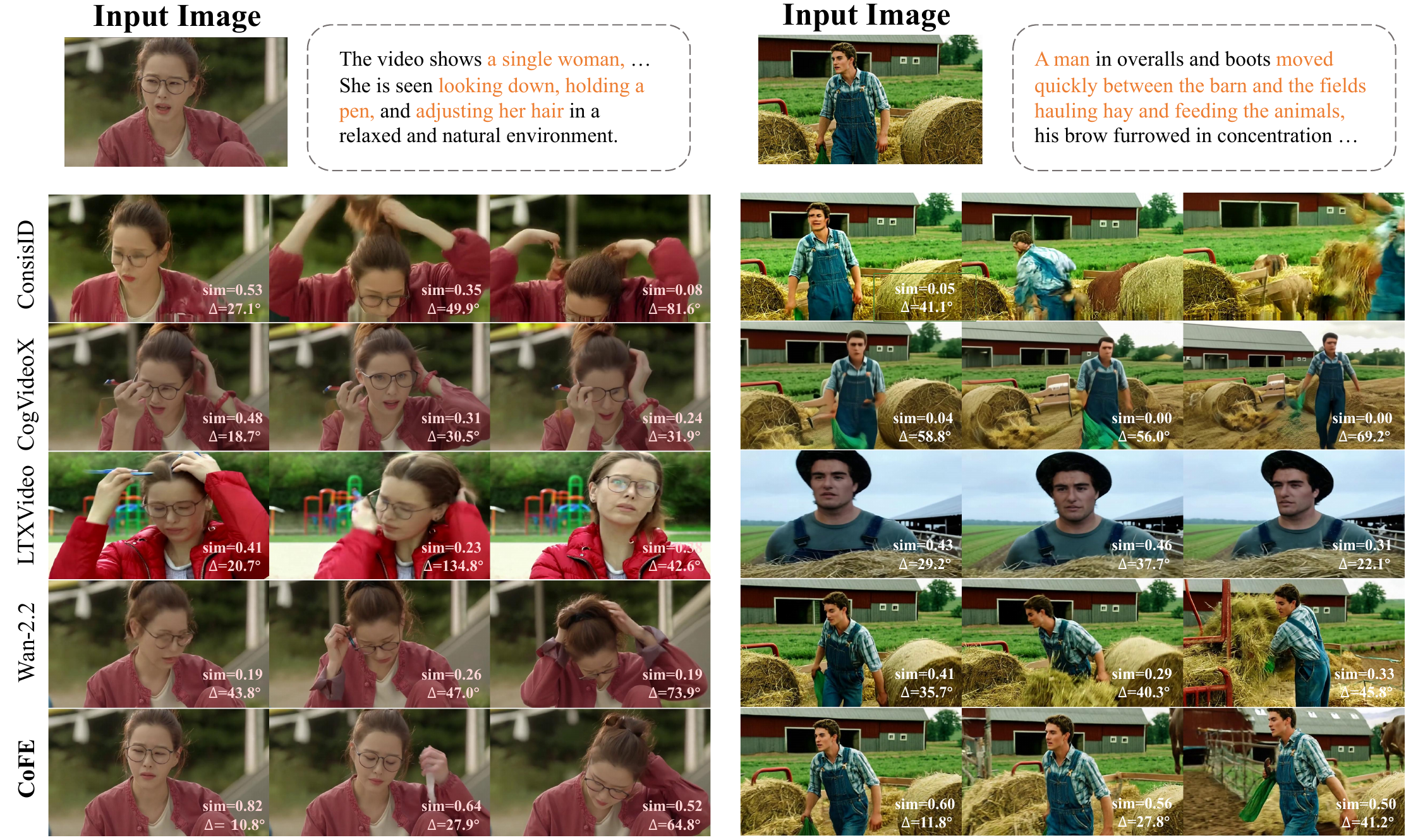}
    \caption{Qualitative comparison of identity preservation and pose handling. For each sequence, we report face similarity (sim) to the reference (first-frame) identity and face pose deviation ($\Delta$, in degrees), defined as the angular distance between the current frame’s and the first frame’s 3D face orientations. \method~achieves superior identity preservation and visual quality over existing methods. }
    \label{fig:main_exp}
    \vspace{-5mm}
\end{figure*}
% \footnotetext{The input image comes from the OpenHumanVid dataset~\cite{li2025openhumanvid}.}

%% file: tables/abla_1.tex
\begin{table}[tbp]
\centering
\caption{Comparison of face feature injection methods on \ourdata-Bench. \method~achieves the best identity preservation while maintaining strong text relevance and fidelity.}
\resizebox{\linewidth}{!}{
\begin{tabular}{lcccc}
\hline
\hline
Method & FaceSim-Arc↑ & FaceSim-Cur↑ & CLIP-Score↑ & FID↓ \\ \hline
Wan-I2V~\cite{wan2025wan} & 0.654 & 0.656 & 29.200 & 72.322 \\
w/ Identity Expert & 0.661 & 0.665 & \textbf{29.402} & 80.305 \\
w/ Semantic Expert & 0.681 & 0.684 & 29.181 & 73.702 \\
w/ Detail Expert & 0.672 & 0.675 & 29.079 & 75.467 \\
w/ Concat & 0.732 & 0.735 & 29.143 & \textbf{63.418} \\
w/ \textbf{\method} & \textbf{0.770} & \textbf{0.775} & 29.125 & 63.799 \\ \hline \hline
\end{tabular}
}

\label{tab:abla_1}
\end{table}

%% file: figs/visual_ablation.tex
\begin{figure}[tbp]
    \centering
    \includegraphics[width=0.8\linewidth]{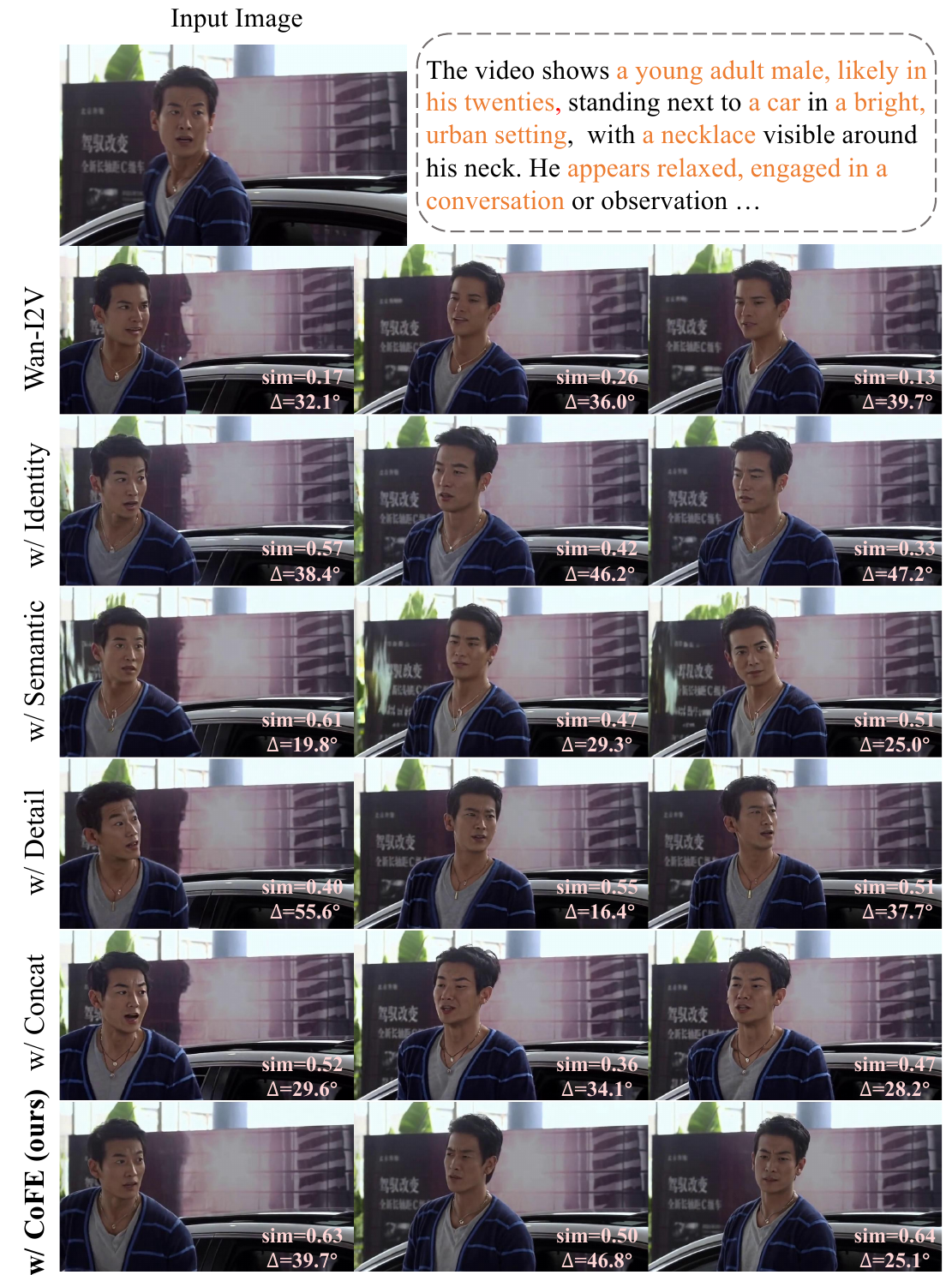}
    \caption{Qualitative comparison between \method~module and other face feature injection methods. Our proposed \method~demonstrates superior identity preservation under large face angle variations in video sequences. }
    \label{fig:abla}
    \vspace{-2mm}
\end{figure}
% The input images are from OpenHumanVid~\cite{li2025openhumanvid}.

%% file: tables/abla_datasets.tex
\begin{table}[tbp]
\centering
\caption{Ablation study on training datasets. Improvements from Pose Filtering and Speech Disambiguation (SD) are evident: \ourdata~w/o SD outperforms OpenHumanVid, and full \ourdata~further boosts identity fidelity and semantic alignment.}
\resizebox{\linewidth}{!}{
\begin{tabular}{lcccc}
\hline \hline
Dataset & FaceSim-Arc↑ & FaceSim-Cur↑ & CLIP-Score↑ & FID↓ \\ \hline
OpenHumanVid & 0.801 & 0.813 & 28.659 & 75.578 \\
\ourdata~w/o SD & 0.807 & 0.816 & 28.634 & 73.128 \\
\ourdata  &\textbf{0.814} & \textbf{0.827} & \textbf{28.729} & \textbf{70.727} \\
 \hline \hline
\end{tabular}
}
\vspace{-4mm}
\label{tab:abla_datasets}
\end{table}

%% file: secs/6_conclusion.tex
\section{Conclusion}
In this paper, we address facial identity preservation under large pose variations in image-to-video generation through two key contributions.  
First, we introduce the Collaborative Face Experts Fusion (\method) module, which dynamically fuses identity-sensitive, semantic, and detail-oriented face features through a layer-adaptive cross-attention gating mechanism. Second, we curate \ourdata-180K through a three-stage pipeline: facial constraints filtering for large-pose diversity, identity consistency analysis for cross-frame identity stability, and speech disambiguation to align captions with speaking status, effectively reducing lip motion and text ambiguity.
To support rigorous evaluation in the challenging regime of large face poses, we introduce \ourdata-Bench, a new benchmark specifically designed for this scenario. Experiments show that \method~achieves state-of-the-art performance on both the general-purpose VIP-Test~\cite{vip-200k} and our specialized \ourdata-Bench, demonstrating consistent improvements in identity preservation and overall video quality across diverse generation settings.

% Limitations放到suppl写吧
% \noindent \textbf{Limitations.} Our method is limited to single-person generation and cannot handle multi-person scenes, where identity interference may occur. Future work will extend \method~with mask-based segmentation for person-specific feature injection, and expand \ourdata~180K and its benchmark to include multi-person interactions. 

% \noindent \textbf{Limitations.} A limitation of our current method is that it focuses solely on single-person identity preservation and cannot handle multi-person scenarios, where cross-identity interference may arise. To address this, future work will extend to multi-person scenarios by integrating mask-based segmentation techniques to enable individual-specific feature injection via the \method~module, while also expanding the dataset to include multi-person interaction samples and establishing a corresponding benchmark.